\documentclass{article}
\usepackage{iclr2026_conference,times}
\usepackage[T1]{fontenc}
\usepackage[utf8]{inputenc}
\usepackage{amsmath,amssymb}
\usepackage{graphicx}
\usepackage{booktabs,array,tabularx}
\usepackage[table]{xcolor}
\usepackage{enumitem}
\usepackage[section]{placeins}
\usepackage{url}
\usepackage[colorlinks=true,linkcolor=blue!55!black,citecolor=blue!55!black,urlcolor=blue!55!black]{hyperref}

\title{Self-evolving Agent Harness}
\author{Chuyang Wei\textsuperscript{1,3}, Haoxiang Guan\textsuperscript{1,3}, Minghang Zhu\textsuperscript{1}, Kexin Zhou\textsuperscript{1,3},\\
\textbf{Jiyan He\textsuperscript{1,2}, Xiren Zhou\textsuperscript{3}, Huanhuan Chen\textsuperscript{3}, Yifei Shen\textsuperscript{1,2,\textdagger}}\\[3pt]
\textsuperscript{1}Zhongguancun Academy,
\textsuperscript{2}Zhongguancun Institute of Artificial Intelligence\\
\textsuperscript{3}University of Science and Technology of China\\
\textsuperscript{\textdagger}Corresponding author}
\iclrfinalcopy
\hypersetup{pdftitle={Self-evolving Agent Harness},pdfauthor={Chuyang Wei, Haoxiang Guan, Minghang Zhu, Kexin Zhou, Jiyan He, Xiren Zhou, Huanhuan Chen, Yifei Shen}}

\begin{document}
\maketitle
\lhead{Self-evolving Agent Harness}
\rhead{Preprint}
\raggedbottom

\begin{abstract}
Harness self-evolution revises an agent's software scaffolding, including prompts, memory, workflows, and executable components, while keeping model weights fixed. We formalize a feedback-driven process in which accepted revisions persist across runs. The framework separates update targets from proposal and admission mechanisms. It distinguishes changes to interpreted state from revisions to tools and code, repository-level control logic, and plugin composition. We characterize search through evolutionary optimization and examine its reliability through continual PAC-learning. Under explicit assumptions, the analysis yields an expected generalization bound, a retention bound conditional on historical regression budgets, and a high-probability admission rule using fresh, independent validation data. We examine how persistent harness revisions affect knowledge sharing and organizational workflows, with implications for scientific reliability, physical safety, and institutional accountability. Cases from prior work illustrate model--harness co-design, distributed execution, and plugin reuse. Reliable deployment requires continual validation as task distributions and evaluators change, with independent safeguards for learning, runtime privileges, and consequential actions. \textbf{We provide a comprehensive 66-page discussion of related work, theory, algorithms, evaluation, applications, and the broader impact of Harness RSI, together with continuously updated code and supplementary materials, at https://github.com/zgcagi/Harness-RSI.}

\end{abstract}

\begin{figure}[!ht]
\centering
\includegraphics[width=\linewidth,trim=0.6bp 0.6bp 0.6bp 0.6bp,clip]{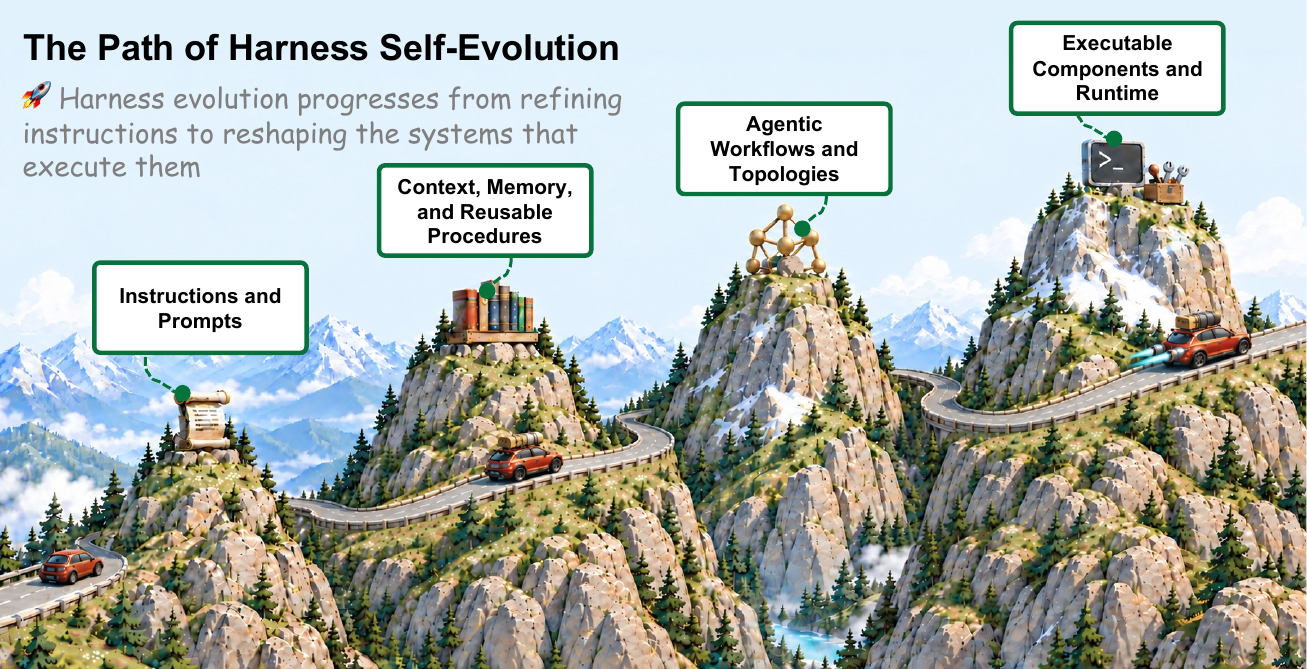}
\caption{From reusable instructions and procedures to revisable executable components.}
\label{fig:harness-path}
\end{figure}

\section{Introduction}\label{sec:intro}

As tasks and environments change after deployment, agents need mechanisms for continued adaptation. Harness self-evolution uses execution feedback to revise instructions, memory, and control logic while keeping model weights fixed. Later runs load accepted revisions, and the update mechanism remains available to learn from subsequent experience (Section~\ref{sec:definition}). This process builds on recursive self-improvement and self-referential problem solving in the context of contemporary agents~\citep{good1965ultraintelligent,schmidhuber2003godel}.

Adaptation can revise model weights or the harness governing prompts, memory, workflows, and tools. Our focus on the latter rests on three considerations. First, the harness directly shapes a fixed model's behavior, while reusable procedures preserve task constraints and lessons from earlier failures~\citep{chan2025mlebench,zhao2024expel,wang2024voyager}. Second, harness revisions reuse an existing model deployment, avoiding retraining and additional weight checkpoints. This can reduce adaptation and maintenance costs and permits updates through proprietary APIs without access to model weights. Third, instructions, memory records, and code support direct inspection, version comparison, and change tracking; attributing behavioral differences between model checkpoints can require additional controlled experiments. We hold the task model fixed to isolate harness self-evolution, with joint model--harness evolution addressed in Section~\ref{sec:future-coevolve}.

Existing methods vary in what is modified, how candidates are generated, and how revisions are admitted~\citep{shinn2023reflexion,zhao2024expel,wang2024voyager,zhang2025aflow,robeyns2025selfimproving}. We first classify the \emph{update target}: the state modified and inherited by later runs. Interpreted-state updates revise instructions, stored experience, or task-level coordination over predefined capabilities; executable revisions change capability implementations, runtime control logic, or plugin-based service composition (Section~\ref{sec:harness}; Figure~\ref{fig:harness-path}). Empirical selection alone does not establish generalization or retention. Section~\ref{sec:analysis} relates candidate search to evolutionary structures, then uses stability and concentration results to state conditions for expected generalization, bounded accumulated regressions, and high-probability admission. Persistent state also determines how revisions propagate and which consequences software recovery can address, motivating system design choices, continual validation, and independent safety controls (Sections~\ref{sec:social}--\ref{sec:future}).

This work makes three contributions:
\begin{itemize}[leftmargin=*,itemsep=3pt,topsep=3pt]
\item \textbf{Definition and Operational Taxonomy (Sections~\ref{sec:definition}--\ref{sec:harness}).} We define a persistent feedback-driven lifecycle and organize interpreted-state updates and executable revisions by their mutation boundaries.
\item \textbf{Conditional Theoretical Analysis (Section~\ref{sec:analysis}).} We map harness search to evolutionary structures and derive conditions for expected generalization, bounded accumulated regressions, and high-probability admission from classical stability and concentration results.
\item \textbf{Systems Implications and Design Trade-offs (Sections~\ref{sec:social}--\ref{sec:future}).} We analyze propagation and repair burdens in five settings and connect attributed cases (Sections~\ref{sec:harness} and~\ref{sec:future}) to model--harness pairing, deployment, and reuse decisions, with continual validation and safety constraints.
\end{itemize}

Associated resources are available at \url{https://github.com/zgcagi/Harness-RSI}.

\section{Definition of Harness Self-Evolution}\label{sec:definition}

\paragraph{Harness.}
Following the model--harness decomposition~\citep{deepseek2026harness}, we define a harness $H$ as the software and configuration governing a model $M$'s execution in a task environment $E$. Table~\ref{tab:harness-components} summarizes seven runtime functions. Model adapters are part of $H$; learned parameters are part of $M$. Execution produces a rollout, which is scored by $J$:
\begin{equation}
\tau\sim\operatorname{Run}(M,H,E),\qquad y=J(\tau).
\end{equation}
The final assessment $J$ includes scoring, outcome verification, and the records that determine the result. Candidates may revise internal critics or provisional checks within $H$, but cannot modify $J$.

\begin{table}[!ht]
\vspace{-6pt}
\centering
\caption{Runtime functions of a harness. An implementation may combine these functions rather than assign each to a separate module.}
\label{tab:harness-components}
\small
\definecolor{harnessblue}{HTML}{2563EB}
\setlength{\tabcolsep}{3pt}
\renewcommand{\arraystretch}{0.98}
\begin{tabularx}{\linewidth}{@{}>{\raggedright\arraybackslash}p{0.22\linewidth}>{\raggedright\arraybackslash}p{0.28\linewidth}>{\raggedright\arraybackslash}X@{}}
\toprule
\rowcolor{harnessblue!10}
\textbf{Component} & \textbf{Runtime role} & \textbf{Mechanisms}\\
\midrule
\rowcolor{harnessblue!4}
Agent loop and execution control & Controls execution and termination & Model and tool calls; continuation, stopping, cancellation, and recovery\\
Tool and skill orchestration & Exposes callable capabilities and procedures & Interfaces and registration; dispatch; skill loading and invocation\\
\rowcolor{harnessblue!4}
Context construction & Prepares each model input & System instructions, tool descriptions, dialogue history, and added context\\
Memory and state management & Preserves information across calls or sessions & Retrieval and reusable memory; persistence, resumption, and forking\\
\rowcolor{harnessblue!4}
Context compaction & Keeps model inputs within context limits & Summarization; retention and trigger policies; pruning tool outputs\\
Scheduling and coordination & Assigns work across execution loops & Delegation to subagents; task scheduling; background work\\
\rowcolor{harnessblue!4}
Execution infrastructure & Enforces boundaries and records activity & Sandboxes, permissions, approvals, storage, model adapters, logs, and replay\\
\bottomrule
\end{tabularx}
\vspace{-6pt}
\end{table}

\paragraph{Harness self-evolution.}
Harness self-evolution uses execution experience to iteratively revise $H$, aiming to improve task performance or resource efficiency while holding $M$ fixed. At round $k\ge1$, the incumbent $H_{k-1}$ generates a batch of $b$ rollouts, each including its trajectory, outcome, and available feedback:
\begin{align}
\mathcal B_k&=\{\tau_{k,i}\}_{i=1}^{b},\qquad
\tau_{k,i}\sim\operatorname{Run}(M,H_{k-1},E_{k,i}),
\label{eq:harness-rollout-batch}\\
H_k&=\operatorname{Update}(H_{k-1},\mathcal B_k).
\label{eq:harness-update}
\end{align}
The update operator combines diagnosis, proposal, assessment, and admission. It may draw on historical feedback, archived candidates, and additional validation rollouts. If no revision is admitted, the incumbent is retained: $H_k=H_{k-1}$.

Each retained revision changes the state used for subsequent execution, whose feedback can guide the next update. This lifecycle requires:
\begin{enumerate}[leftmargin=*,topsep=3pt,itemsep=3pt]
\item \textbf{Operational feedback and update continuity.} The system uses feedback from its own executions to construct or select revisions algorithmically, and retains the mechanism for further updates. Environments, automated assessors, and people may provide feedback; admission may require human approval.
\item \textbf{Persistence across runs.} Accepted revisions are stored and loaded in a subsequent run, allowing experience from one interaction to influence later execution.
\end{enumerate}
These requirements concern repeatability: one observed update can demonstrate the capability if both hold. Appending logs or compacting context under fixed rules does not revise the harness; learning reusable procedures or changing memory policies can. Revising the update mechanism adds another level of recursion.

\paragraph{Scope and terminology.}
\emph{Self-evolution} denotes the update process; \emph{self-improvement} denotes a gain established by comparison. We use \emph{harness RSI} (recursive self-improvement) as the name of the research area, without assuming that each revision helps. Weight-only adaptation and scaffolds without an update mechanism fall outside this scope. When a system also trains weights, we examine its harness updates with the task model held fixed.

\section{The Taxonomy of Harness Self-Evolution}\label{sec:harness}

EURISKO's heuristic revision and the G\"odel Machine's proof-governed self-rewriting established distinct antecedents of harness self-evolution~\citep{lenat1983eurisko,schmidhuber2003godel}. Contemporary methods use empirical feedback to revise prompts, consolidate persistent memory, and search workflow specifications around a fixed model~\citep{agrawal2026gepa,fernando2024promptbreeder,zhao2024expel,wang2025workflow,zhang2025aflow}. The interpreted/executable distinction concerns the edited artifact's role, not its programming language. Workflow revisions remain interpreted when they change task-level sequencing or routing through fixed coordination interfaces over predefined operators. This includes code that only composes those operators; revising operator implementations or the runtime mechanisms that enact coordination constitutes an executable revision, even under an unchanged executor.

We organize executable targets into component implementations, repository-level control logic, and plugin composition. The first two describe mutation scope; the third describes how components and runtime services are bound and replaced. Plugin composition can span component implementations and repository-level logic, so a revision may belong to more than one category. Classifying the update target does not determine whether a method satisfies the feedback, continuity, and persistence requirements in Section~\ref{sec:definition}.

\subsection{Tools and Code Patches}\label{sec:objects-runtime}

Component-level revision confines mutation to callable procedures or designated modules while keeping surrounding orchestration outside the editable surface. Where entry points and I/O contracts remain fixed, candidates replace component implementations; tool synthesis may instead extend the callable interface. Skill libraries retain executable functions for later retrieval and composition~\citep{wang2024voyager,yu2025polyskill}; patch-based architectures, including Self-Harness, HarnessCompass, and Agentic Harness Engineering, constrain trace-guided mutation to selected components~\citep{zhang2026selfharness,zhang2026harnesscompass,lin2026agentic}. Editing memory-management functions changes the harness machinery, whereas curating stored entries changes its interpreted state~\citep{zhang2025memevolve}. Within this boundary, candidate archives preserve alternatives for subsequent search~\citep{luo2026harnessbank}, while protocol perturbations test whether an evolved harness's gain over the initial harness survives changes in assessment conditions~\citep{zhu2026badgenius}. Stable interfaces facilitate replacement and local checks, but changed outputs or shared state can propagate regressions beyond the edited module.

\subsection{Repository and Runtime Logic}\label{sec:objects-repo}

Repository-level search exposes context construction, orchestration loops, execution hooks, recovery, and termination to revision. The boundary concerns control over interactions among harness components, rather than the number of files edited. Whole-agent synthesis searches complete control programs, while repository evolution retains implementation lineages for later selection and revision~\citep{hu2025automated,lee2026metaharness,robeyns2025selfimproving,zhang2026darwin,zhang2026darwinx,wang2025huxley}. Changes to shared context, scheduling, or recovery can alter the inputs and execution assumptions of otherwise unchanged components. This coupling motivates edit constraints, transfer-oriented admission, and deployment checks~\citep{xia2026rrsi,cai2026moss}. Archive membership preserves a candidate for further search; adoption additionally requires promoting the selected program into the runtime used by later tasks. Extending mutation to the optimizer's prompts, hooks, and tools makes the update mechanism itself evolvable~\citep{wang2026verse}. An exported program satisfies the continuing lifecycle only if its deployment retains an update mechanism.

HarnessOpt extends the SkillOpt-Lite study from skill editing to runtime revision, improving spreadsheet previews, output checks, and recovery after tool failures~\citep{shen2026skilloptlite}. On SpreadsheetBench (Table~\ref{tab:skillopt-case}), GPT-5.4-nano with HarnessOpt and skill reaches 77.58\%, exceeding GPT-5.5 with SkillOpt (76.20\%); across all four models, HarnessOpt with skill improves on SkillOpt-Lite by 5.69--11.39 percentage points. This crossover illustrates model--harness pairing, but does not establish superiority at matched cost.

\begingroup
\setlength{\intextsep}{7pt}
\begin{table}[!ht]
\centering
\caption{SpreadsheetBench accuracy (\%), converted from proportions in Table~3 of \citet{shen2026skilloptlite}. HO: HarnessOpt; + skill: with-skill condition.}
\label{tab:skillopt-case}
\small
\setlength{\tabcolsep}{4pt}
\begin{tabular}{@{}lrrrrr@{}}
\toprule
Model & Initial & SkillOpt & SkillOpt-Lite & HO only & HO + skill\\
\midrule
GPT-5.4-nano & 29.89 & 51.60 & 66.19 & 76.51 & \textbf{77.58}\\
GPT-5.4-mini & 28.21 & 47.50 & 73.31 & 81.14 & \textbf{82.56}\\
GPT-5.4      & 39.86 & 61.50 & 79.36 & 83.63 & \textbf{85.05}\\
GPT-5.5      & 37.37 & 76.20 & 79.72 & 83.63 & \textbf{85.77}\\
\bottomrule
\end{tabular}
\end{table}
\endgroup

\subsection{Everything Is a Plugin}\label{sec:objects-plugin}

Plugin architectures organize mutation through declared interfaces, dependency constraints, and lifecycle operations. The editable object may be a component implementation or a binding that determines which implementation supplies a runtime service, such as memory storage, tool dispatch, or the agent loop. Composition here therefore reaches beyond task-level routing among fixed operators. Kernel-defined substrates specify how these services are connected, mounted, and replaced~\citep{deepseek2026harness,shi2026programming}; feedback-driven systems revise typed primitives, versioned resources, or scoped modules under interface and validation constraints~\citep{chen2026harnessx,li2026argus,zhang2026autogenesis,wu2026modularrsi}. This organization can expose both local capabilities and core runtime services while retaining explicit constraints on their composition. Everything Is a Plugin (EIP) extends it to capabilities generated, validated, and mounted during operation. Interface contracts support compatibility checks and component replacement; they do not establish behavioral isolation or safe admission. Adoption still requires validation of the composed behavior.

Using DSH Everything, a Codex plugin lets the agent generate tools during execution and retain them for later tasks~\citep{shen2026dsheverything,pan2026eip}. In a simulated office workload, spreadsheets from different company regions vary in language and content but share similar calculations. Codex creates batch-lookup and summarization tools: first-task tool calls fall from 13 to 3, while wall time rises by 54.9\% and tokens by 15.7\%. A separately reported cumulative series shows savings from the second task, reaching 110,351 tokens and 22.97~s after five tasks (Table~\ref{tab:eip-case}). The first-task totals imply increases of 6,453 tokens and 8.51~s, whereas cumulative row~1 reports 4,895 tokens and 6.74~s. This unresolved difference prevents treating the two reports as a single cost trajectory; negative cumulative differences indicate savings.

\begin{table}[!ht]
\centering
\caption{EIP spreadsheet costs~\citep{pan2026eip}.}
\label{tab:eip-case}
\small
\setlength{\tabcolsep}{4pt}
\renewcommand{\arraystretch}{1.12}
\begin{minipage}[t]{0.47\linewidth}
\vspace{0pt}
\textbf{(a) First task}\par\smallskip
\renewcommand{\arraystretch}{1.68}
\begin{tabularx}{\linewidth}{@{}Xrr@{}}
\toprule
Metric & Baseline & EIP Aggregate\\
\midrule
Tool calls & 13 & 3\\
Tokens & 41,119 & 47,572\\
Wall time (s) & 15.49 & 24.00\\
\bottomrule
\end{tabularx}
\end{minipage}\hfill
\begin{minipage}[t]{0.47\linewidth}
\vspace{0pt}
\textbf{(b) Cumulative EIP $-$ baseline}\par\smallskip
\begin{tabularx}{\linewidth}{@{}Xrr@{}}
\toprule
Tasks & $\Delta$ tokens & $\Delta$ wall time (s)\\
\midrule
1 & $+4{,}895$ & $+6.74$\\
2 & $-19{,}907$ & $-3.97$\\
3 & $-47{,}375$ & $-8.44$\\
4 & $-77{,}542$ & $-13.87$\\
5 & $-110{,}351$ & $-22.97$\\
\bottomrule
\end{tabularx}
\end{minipage}
\end{table}
\FloatBarrier

\begin{table}[!ht]
\centering
\caption{LiveBrowseComp results~\citep{pan2026eip}.}
\label{tab:eip-browse}
\begin{minipage}{0.76\linewidth}
\small
\setlength{\tabcolsep}{6pt}
\renewcommand{\arraystretch}{1.12}
\begin{tabularx}{\linewidth}{@{}Xr@{}}
\toprule
\textbf{Model / harness} & \textbf{Score}\\
\midrule
\rowcolor{black!5}[0pt][0pt]
\multicolumn{2}{@{}l@{}}{\emph{$<40$B}}\\
\textbf{XYZ + dsh-everything} & \textbf{51.39}\\
XYZ-Aquila-mini & 48.7\\
Nex-N2-mini & 41.4\\
MiroThinker 1.7 mini & 34.9\\
apodex-mini & 32.8\\
Agents-A1 & 29.6\\
\addlinespace[3pt]
\rowcolor{black!5}[0pt][0pt]
\multicolumn{2}{@{}l@{}}{\emph{$<400$B}}\\
XYZ-Aquila-pro & 53.7\\
Nex-N2-Pro & 50.4\\
MiroThinker 1.7 & 34.1\\
\addlinespace[3pt]
\rowcolor{black!5}[0pt][0pt]
\multicolumn{2}{@{}l@{}}{\emph{$>1$T or closed-source}}\\
DeepSeek V4-Pro & 38.3\\
Kimi-K2.6 & 31.7\\
\bottomrule
\end{tabularx}
\end{minipage}
\end{table}
\FloatBarrier
\newpage

The same source reports a LiveBrowseComp score of 51.39 for XYZ + dsh-everything, compared with 48.7 for XYZ-Aquila-mini, placing it first among the six displayed $<40$B entries and second among all eleven (Table~\ref{tab:eip-browse}). Table~\ref{tab:eip-browse} preserves the source's model groups and scores, whose unit is unspecified. These reports illustrate potential reuse savings and model--harness pairing. Repeated runs, matched budgets, and matched task quality are needed to establish causal plugin gains and general reuse benefits.

\section{Theory of Harness Self-Evolution}\label{sec:analysis}

Many harness optimizers explore discrete instructions, skills, and code through execution-guided proposals rather than gradients through the edits. Evolutionary algorithms (EAs) characterize the search structure; continual-learning analysis relates empirical gains to generalization and capability retention. Recent work likewise distinguishes optimization from generalization and links reliable adoption to verification cost~\citep{gan2026harnessevolutionlearning,cai2026reliableharness}.

\subsection{An Evolutionary Perspective on Harness Search}\label{sec:ea-correspondence}

The EA view separates four aspects of an update: the representation of mutable state, the variation that generates edits, the assessment of their fitness, and the selection of state to retain. Traces and critiques can guide variation, while admission depends on the resulting assessment.

\begin{figure}[!t]
\centering
\includegraphics[width=\linewidth]{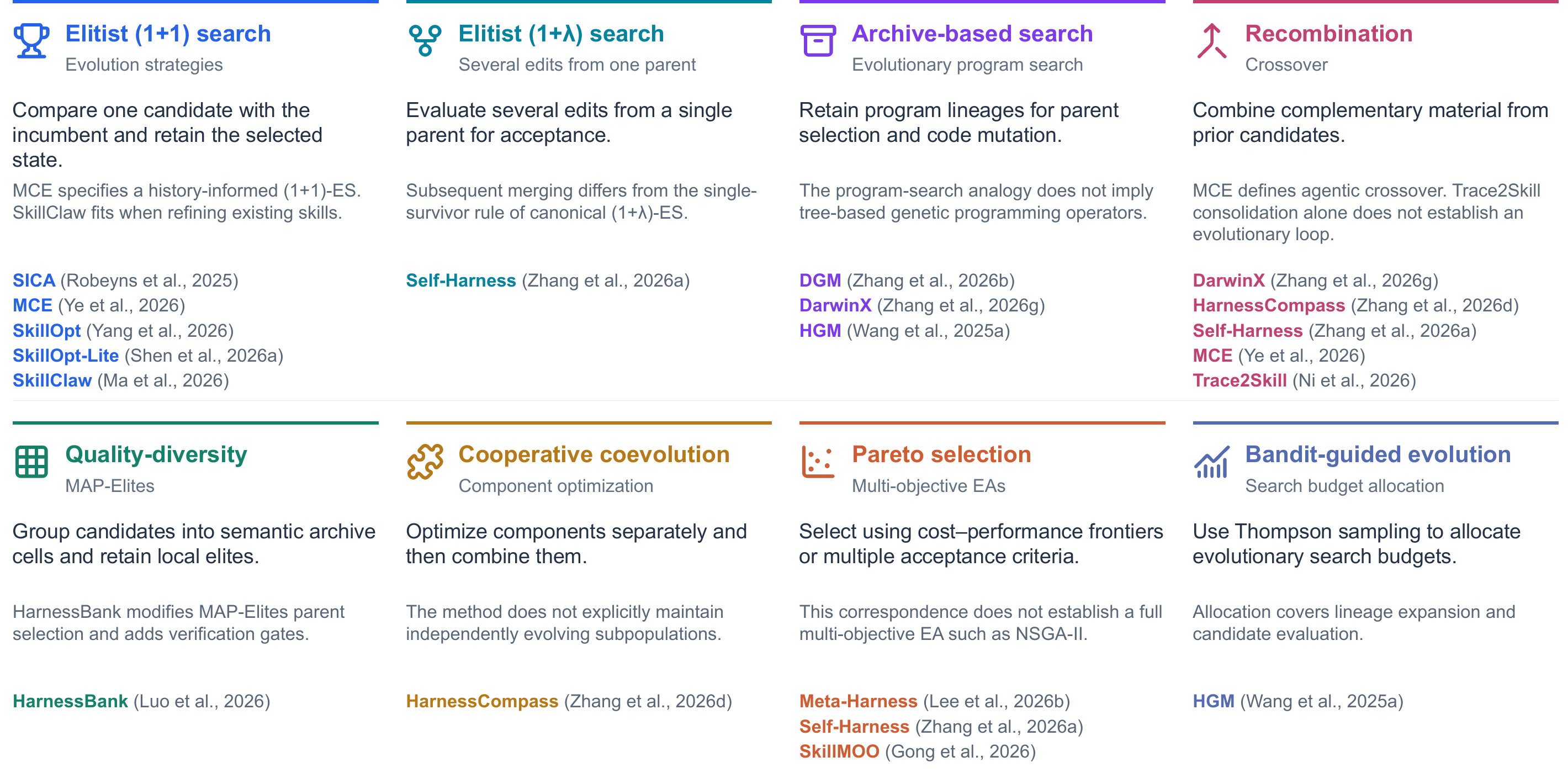}
\caption{Evolutionary algorithm correspondences in harness and skill optimization. Each panel summarizes a search structure, the basis and limits of its correspondence, and representative methods. The categories overlap: a method may instantiate more than one structure, and a correspondence does not imply equivalence to the full canonical algorithm.}
\label{fig:ea-harness-skill}
\end{figure}

\emph{Local elitism} concentrates search around an incumbent; \emph{archive-based diversity} preserves alternative lineages; \emph{Pareto selection} balances competing objectives. Figure~\ref{fig:ea-harness-skill} summarizes eight algorithmic correspondences, their limits, and representative methods. Because all three mechanisms select on empirical fitness, accepted revisions can overfit or forget, motivating the generalization and retention analysis below.

\nocite{robeyns2025selfimproving,ye2026meta,yang2026skillopt,shen2026skilloptlite,ma2026skillclaw,zhang2026selfharness,zhang2026darwin,zhang2026darwinx,wang2025huxley,zhang2026harnesscompass,ni2026trace2skill,luo2026harnessbank,lee2026metaharness,gong2026skillmoo}

\subsection{Generalization and Retention from a Continual PAC-Learning Perspective}\label{sec:generalization}

Lasting capability requires controls on generalization, forgetting, and admission uncertainty. Continual PAC-learning studies retention under memory constraints~\citep{chen2022memory}; here we use stability and concentration under explicit sampling assumptions.

\paragraph{Evolution state and sampling.}
Let $X_k$ record the harness, skills, archive, and feedback memory after round $k$, with $H_k=\operatorname{Deploy}(X_k)$ and $X_k=\mathcal A_k(X_{k-1},D_k;U_k)$. For stability, assume $D_k$ contains $n_k$ tasks drawn from an exogenous distribution $\mathcal D_k$, independently within and across rounds. Auxiliary randomness and external assessment samples in $U_k$ are independent of these draws; task-derived feedback remains part of the update process. Fix the model and evaluator, and let $\ell(H,z)\in[0,1]$ denote loss averaged over execution randomness. Define
\begin{equation}
R_s(H)=\mathbb E_{z\sim\mathcal D_s}\ell(H,z),\qquad
\widehat R_s(H)=\frac1{n_s}\sum_{i=1}^{n_s}\ell(H,z_{s,i}).
\end{equation}

\paragraph{Stability under task replacement.}
Algorithmic stability links task sensitivity to generalization~\citep{shalev2010learnability}. Replace $z_{s,i}$ by an independent draw from $\mathcal D_s$ and rerun all downstream updates, including feedback, archives, and selection, under coupled randomness to obtain $H_k^{(s,i)}$. For $s\le k$, define $\beta_{s,k}=\frac1{n_s}\sum_{i=1}^{n_s}\mathbb E\!\left[\left|\ell(H_k,z_{s,i})-\ell(H_k^{(s,i)},z_{s,i})\right|\right]$. Replace-one exchangeability yields
\begin{equation}
\left|\mathbb E[R_s(H_k)-\widehat R_s(H_k)]\right|\le\beta_{s,k}.
\label{eq:harness-stability}
\end{equation}
The bound controls the magnitude of the expected generalization gap, not the gap in any individual run. Small code edits do not by themselves imply stability of the full update process.

\paragraph{Retention and accumulated regressions.}
Write $F_{s,k}=R_s(H_k)-R_s(H_s)$, with positive values indicating forgetting and negative values backward transfer. For fixed regression budgets $\varepsilon_{s,u}$, suppose each later transition satisfies $R_s(H_u)-R_s(H_{u-1})\le\varepsilon_{s,u}$. Telescoping gives $F_{s,k}\le\sum_{u=s+1}^{k}\varepsilon_{s,u}$; combined with stability at round $s$, this yields
\begin{equation}
\mathbb E R_s(H_k)\le\mathbb E\widehat R_s(H_s)+\beta_{s,s}
+\sum_{u=s+1}^{k}\varepsilon_{s,u}.
\label{eq:harness-forgetting}
\end{equation}
The term $\beta_{s,s}$ measures sensitivity of the complete round-$s$ update. The historical-risk budgets are separate assumptions, not implications of stability, and their allowances accumulate. Checking them requires groupwise uncertainty control for the deployed composition, including merged edits and shared routers.

\paragraph{Admission with fresh validation.}
Let $\bar R_k(H)=k^{-1}\sum_{s=1}^{k}R_s(H)$ denote risk on the uniform historical mixture. At each of $T$ rounds, fix at most $K$ candidates, then draw a fresh validation set $V_k$ of $m$ independent tasks from this mixture, independent of candidate construction conditional on the prior history. For candidate $H'$, define
\begin{equation}
\begin{aligned}
\Delta_k(H')&=\bar R_k(H')-\bar R_k(H_{k-1}),\\
\widehat\Delta_k(H')&=\frac1m\sum_{z\in V_k}
[\ell(H',z)-\ell(H_{k-1},z)].
\end{aligned}
\end{equation}
Paired differences lie in $[-1,1]$, with range width $2$. Conditional Hoeffding bounds followed by a union bound over at most $KT$ comparisons give, simultaneously with probability at least $1-\delta$~\citep{hoeffding1963probability},
\begin{equation}
\Delta_k(H')\le\widehat\Delta_k(H')+
\sqrt{\frac{2\ln(KT/\delta)}{m}}.
\label{eq:harness-paired-gate}
\end{equation}
Accept only candidates with negative upper bounds. Edits or merges informed by $V_k$ need fresh validation; mixture gains do not ensure groupwise retention. HarnessBank reuses training tasks for screening before a separate final test; Self-Harness reuses a held-out regression split~\citep{luo2026harnessbank,zhang2026selfharness}. These search-time checks lack the assumed independence. Secrecy and comparison-count corrections do not remove adaptive dependence; use fresh samples or adaptive-valid inference~\citep{dwork2015holdout}.

\paragraph{Scope of the guarantees.}
Independent tasks determine the sample size; additional rollouts can reduce execution noise without adding tasks. Equation~\eqref{eq:harness-paired-gate} also covers unbiased paired-loss estimates in $[-1,1]$ from independent task clusters. Correlated sessions need grouping or dependence-aware analysis, and non-additive metrics require separate concentration arguments. Admission is certified on the historical mixture, not arbitrary future distributions. If evaluator and target risk differ by at most $\eta$ for each compared harness, a sufficient condition for target-risk improvement is
\begin{equation}
\widehat\Delta_k(H')+\sqrt{\frac{2\ln(KT/\delta)}{m}}+2\eta<0.
\label{eq:harness-target-gate}
\end{equation}
The discrepancy bound requires evidence from executable checks or independent human assessment. Without such a bound, the certificate concerns only evaluator-defined loss. Section~\ref{sec:future-eval} discusses continuing validation as task distributions and evaluators change.

\section{Applications and Societal Impacts}\label{sec:social}\label{sec:applications}

Persistent harness revisions preserve learned procedures across tasks. We examine their implications for knowledge sharing, reliability, and accountability in five application settings.

\paragraph{Personal workspaces.}
Revisable instructions, memory, and skills preserve project conventions and user preferences across programming, writing, and research sessions~\citep{anthropic2026claudecodememory,github2026copilotmemory,anthropic2025agentskills}. Playbook revision and trajectory distillation convert execution feedback into reusable procedures that can reduce repeated setup and diagnosis~\citep{zhang2026agentic,ni2026trace2skill,shen2026skilloptlite}. As updates accumulate, users may struggle to track changes: temporary workarounds become standing rules, and injected instructions persist across tasks~\citep{gadgil2026bad}. Even successful task executions can yield skills that generalize harmfully~\citep{zhang2026skillpoison}. Reviewable changes and artifact provenance support ongoing curation; recovery must encompass derived skills, memory, and caches as well as tracked code.

\paragraph{Enterprise automation.}
Shared skill libraries and revisable agent components let teams reuse incident diagnoses, API adaptations, and recovery procedures as operational requirements change~\citep{ma2026skillclaw,liu2026skillforge,zhang2026darwinx,lin2026agentic,cai2026moss}. Local workarounds can spread to teams with different permission or data assumptions. Shared faulty procedures may cause correlated errors that escape detection, extending concerns about algorithmic monoculture to persistent harness artifacts~\citep{kleinberg2021algorithmic,bommasani2022picking}. Remediation requires identifying consumers, withdrawing affected versions, restoring dependent workflows, and investigating unauthorized actions. Version-linked deployment records and staged rollout support containment, while service disruption or data exposure may require organizational redress after software recovery.

\paragraph{Scientific discovery.}
Persistent settings, failure reports, and search heuristics help collaborators reuse earlier trials~\citep{qu2026coral,li2026argus}. Diagnostic evaluations also document failures to check claims against evidence and artifacts~\citep{fei2026autoresearch}. When evolving protocols exploit provisional checks or imperfect external metrics, specification gaming can reinforce procedures that fail independent measurement~\citep{skalse2022defining,baker2025monitoring}. These procedures can spread through collaboration or wet-lab transfer, consuming instrument time, reagents, and expert effort before detection. Final assessment $J$ must remain outside the editable surface (Section~\ref{sec:definition}); independent measurement and replication are also needed to establish scientific validity.

\paragraph{Robotics and embodied interaction.}
EMHO revises planning, observation handling, and recovery around a frozen model~\citep{lee2026emho}; RoboRSI revises and reuses a hierarchy of robot skills~\citep{noematrix2026roborsi}. Memory Transfer Planning adapts code from a fixed memory, including transfer from simulation to hardware~\citep{kagaya2025memory}; EmbodiedSWE learns weights from verified simulation trajectories~\citep{shen2026embodiedswe}. Errors in accepted action code can persist across motions, causing excessive force, unsafe recovery, or incorrect contact behavior. Software rollback changes future commands but cannot reverse deformation, wear, or injury. For physical deployment, we recommend keeping emergency stops, force and speed limits, and authorization checks outside the editable surface, with independent action checks and staged trials. Revision-linked command records support investigation and remediation.

\paragraph{Public services and clinical systems.}
Revisable triage, eligibility routing, and case-processing procedures offer an inspectable way to incorporate changing guidelines and operational feedback; current evidence includes prototype clinical interactions and context adaptation on diagnostic and legal tasks~\citep{jiang2025medagentbench,schmidgall2024agentclinic,ye2026meta}. Qualified professionals must validate and authorize consequential revisions. Erroneous preferences can otherwise become persistent rules that reproduce institutional disadvantage across cases~\citep{citron2007technological,shao2026misevolve}. Patients and applicants may lack access to an alternative institution. Software rollback cannot recover a missed treatment window or, by itself, correct denied entitlements and corrupted records. Execution-log inspection aids agent evaluation~\citep{kapoor2026holistic}. For accountability, we recommend recording each decision's rule version, evidence, and admission history, alongside accessible review, appeal, and remediation.

\section{Future Directions}\label{sec:future}

The executable-revision cases in Section~\ref{sec:harness} motivate research on model--harness co-design and plugin lifecycles. We also examine distributed execution, continual validation, and safety boundaries for evolving systems.

\subsection{Model--Harness Co-Design}\label{sec:future-coevolve}

The model--harness crossover in Section~\ref{sec:objects-repo} motivates studying how to allocate adaptation effort between the model and its runtime. Budget-matched studies should compare model selection, skill optimization, runtime revision, and weight updates~\citep{chen2026cotrace,li2026rightfit}, accounting for both adaptation and subsequent execution costs. Testing frozen, transferred, and re-evolved harnesses across checkpoints would distinguish transferable gains from those requiring further harness evolution. Joint adaptation should also examine which update order preserves earlier capabilities as model behavior and runtime assumptions change.

\subsection{Distributed Execution and Edge Infrastructure}\label{sec:future-infrastructure}

Runtime revision (Section~\ref{sec:objects-repo}) can change both execution cost and where components are best placed. A fixed-deployment study illustrates the communication trade-offs: \citet{shen2023autonomousedge} compare a Jetson Nano client, an RTX~4090 edge server, and Azure under a workload of 10\% classification, 40\% captioning, and 50\% visual question answering. Client--edge execution partitions expert networks and compresses task-relevant features; the edge-only baseline transmits losslessly compressed input images.

\begin{figure}[!ht]
\centering
\includegraphics[width=0.5\linewidth]{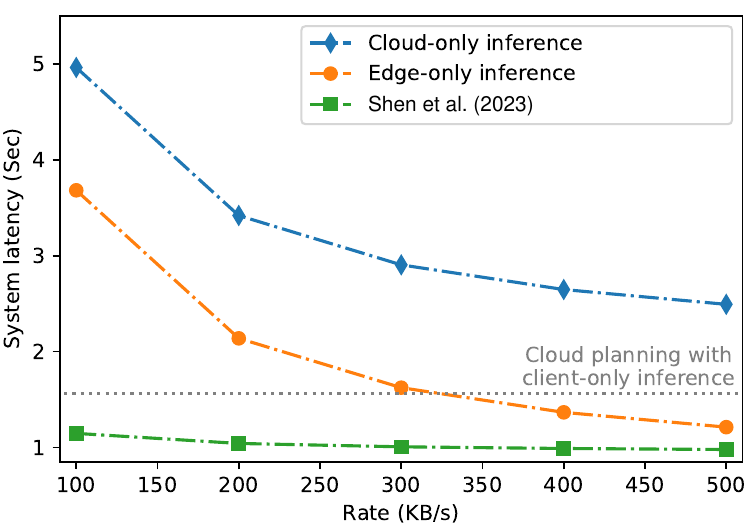}
\caption{Latency adapted from \citet{shen2023autonomousedge}; client--edge in green.}
\label{fig:edge-latency}
\end{figure}

Client--edge co-inference minimizes latency across 100--500~KB/s in this study (Figure~\ref{fig:edge-latency}). Future studies should include harness synthesis, validation, and distribution costs when comparing deployment choices across network conditions and update frequencies, at matched quality and subject to privacy constraints.

\FloatBarrier

\subsection{Plugin Creation and Reuse}\label{sec:future-eip}

The creation overhead and subsequent savings in Section~\ref{sec:objects-plugin} motivate lifecycle-aware decisions about when to reuse, revalidate, regenerate, or retire a plugin. Future studies should measure the number of related tasks needed to recover generation and validation costs at matched quality, including repair and dependency-maintenance costs. Varying task similarity, reuse frequency, and interface changes would reveal when a retained tool remains useful and when regeneration is preferable. Admission should test the resulting composition, since a locally valid plugin can alter shared state or the behavior of other components.

\subsection{Continual Validation under Distribution and Evaluator Drift}\label{sec:future-eval}

Time-stamped tasks~\citep{white2025livebench,jain2024livecodebench,yan2026livemedbench}, version-pinned environments~\citep{merrill2026terminalbench}, and difficulty calibration~\citep{chen2026stresseval} help distinguish harness gains from workload drift. Admission requires fresh independent tasks (Section~\ref{sec:generalization}); adaptive reuse needs separate statistical controls~\citep{dwork2015reusable,blum2015ladder}. Reused calibration tasks cannot replace fresh admission evidence. Judge bias~\citep{lu2025agentrewardbench,wu2024metarewarding} requires independent calibration of $\eta$ in Equation~\eqref{eq:harness-target-gate} after evaluator changes. Protecting $J$ does not prevent specification gaming~\citep{pan2024spontaneous,shao2026misevolve}. Allocating budgets among calibration, admission, and regression checks remains an open problem.

\subsection{Safety Boundaries for Self-Modifying Systems}\label{sec:future-safety}

External safeguards complement model alignment~\citep{bai2022constitutional}. Context-removal and transfer tests can probe harmful skill generalization~\citep{zhang2026skillpoison,wu2026evomal}; data-flow and tool-sequence constraints support runtime enforcement~\citep{doshi2026verifiably}. We recommend least privilege, time-limited grants, and isolation; creating capabilities must not confer authority. Rollback covers only effects within its recovery model~\citep{shi2026programming}, motivating authorization and physical or spending limits for consequential actions. Validating composed behavior and revoking grants as dependencies change remain open problems. Defenses should be evaluated by attack success, isolation time, and legitimate-work overhead.

\section{Conclusion}\label{sec:conclusion}

Our framework defines harness self-evolution through feedback-driven revision, continuing update capability, and persistence around fixed models. It separates mutation boundaries from search and admission, distinguishing expected generalization, budget-constrained retention, and high-probability acceptance under fresh validation.

Persistent revisions preserve knowledge but can spread errors with irreversible consequences. Their deployment requires continual validation under distribution and evaluator drift, with independent controls over learning, runtime privileges, and consequential actions.

\FloatBarrier
\label{last-main-page}
\clearpage
\bibliographystyle{iclr2026_conference}
\bibliography{references}
\clearpage
\appendix
\section{Extended Technical Report}\label{app:extended-report}

This paper presents a self-contained framework for harness self-evolution. For broader literature coverage and extended discussions of evaluation, applications, and societal implications, see our companion technical report, \emph{Harness Self-Evolution: Theory, Algorithms, Evaluation, Applications, and Impact}, at \url{https://openreview.net/pdf?id=m5jsFGXUM2}.

\paragraph{Contents of the companion report.}
\begin{enumerate}[label=\textbf{\arabic*.},leftmargin=1.8em,itemsep=4pt,topsep=6pt,parsep=0pt,partopsep=0pt]
\item \textbf{Introduction}
\item \textbf{Definition of Harness Self-Evolution}
\item \textbf{Interpreted Update Targets}
\begin{enumerate}[label=\arabic{enumi}.\arabic*,leftmargin=2.6em,itemsep=0pt,topsep=1pt,parsep=0pt,partopsep=0pt]
\item From Recursive Self-Improvement to Harness Adaptation
\item Instructions and Prompts
\item Context, Memory, and Reusable Procedures
\item Agentic Workflows and Topologies
\end{enumerate}
\item \textbf{Executable Update Targets}
\begin{enumerate}[label=\arabic{enumi}.\arabic*,leftmargin=2.6em,itemsep=0pt,topsep=1pt,parsep=0pt,partopsep=0pt]
\item Tools and Code Patches
\item Repository and Runtime Logic
\item Everything Is a Plugin
\end{enumerate}
\item \textbf{Theory of Harness Self-Evolution}
\begin{enumerate}[label=\arabic{enumi}.\arabic*,leftmargin=2.6em,itemsep=0pt,topsep=1pt,parsep=0pt,partopsep=0pt]
\item An Evolutionary Perspective on Harness Search
\item Generalization and Retention from a Continual PAC-Learning Perspective
\end{enumerate}
\item \textbf{Evaluating Harness Self-Evolution}
\begin{enumerate}[label=\arabic{enumi}.\arabic*,leftmargin=2.6em,itemsep=0pt,topsep=1pt,parsep=0pt,partopsep=0pt]
\item Four Claims Behind a Task Score
\item Current Evaluation Practices and Benchmarks
\item Longitudinal and Dynamic Evaluation
\end{enumerate}
\item \textbf{Applications of Harness Self-Evolution}
\begin{enumerate}[label=\arabic{enumi}.\arabic*,leftmargin=2.6em,itemsep=0pt,topsep=1pt,parsep=0pt,partopsep=0pt]
\item Personal Assistance and Developer Workspaces
\item Enterprise Automation
\item Scientific Discovery
\item Robotics and Embodied Interaction
\item Public Services and Clinical Systems
\end{enumerate}
\item \textbf{Societal and Economic Implications}
\begin{enumerate}[label=\arabic{enumi}.\arabic*,leftmargin=2.6em,itemsep=0pt,topsep=1pt,parsep=0pt,partopsep=0pt]
\item Productivity, Work, and Education
\item Economic Distribution and Market Structure
\item Innovation and Public-Service Capacity
\item Privacy, Security, and Systemic Stability
\item Governance and Institutional Power
\end{enumerate}
\item \textbf{Future Directions}
\begin{enumerate}[label=\arabic{enumi}.\arabic*,leftmargin=2.6em,itemsep=0pt,topsep=1pt,parsep=0pt,partopsep=0pt]
\item Model--Harness Co-Evolution
\item Optimization Algorithms for Heterogeneous Harness Search
\item Infrastructure for Evolving State
\item Dynamic Evaluation and Benchmark Refresh
\item Safety Boundaries for Self-Modifying Systems
\item Governance, Provenance, and Human Oversight
\end{enumerate}
\item \textbf{Conclusion}
\end{enumerate}

\end{document}